\documentclass[letterpaper]{article} 
\usepackage{aaai2026}  
\usepackage{times}  
\usepackage{helvet}  
\usepackage{courier}  
\usepackage[hyphens]{url}  
\usepackage{graphicx} 
\usepackage{natbib}  
\usepackage{caption} 
\usepackage{algorithm}
\usepackage{algorithmic}
\usepackage{amsmath}
\usepackage{float}
\usepackage{caption}
\usepackage{newfloat}
\usepackage{listings}
\DeclareCaptionStyle{ruled}{labelfont=normalfont,labelsep=colon,strut=off} 
\floatstyle{ruled}
\newfloat{listing}{tb}{lst}{}
\floatname{listing}{Listing}
\usepackage{booktabs}
\usepackage{multirow}
\usepackage{siunitx}
\title{Entropy Guided Diversification and Preference Elicitation in Agentic Recommendation Systems}
\author{
    Dat Tran\equalcontrib\textsuperscript{\rm 1},
    Yongce Li\equalcontrib\textsuperscript{\rm 1},
    Hannah Clay\equalcontrib\textsuperscript{\rm 1},
    Negin Golrezaei\textsuperscript{\rm 2},
    Sajjad Beygi\textsuperscript{\rm 3},
    Amin Saberi\textsuperscript{\rm 1}
}
\affiliations{

    \textsuperscript{\rm 1}Stanford University \\
    \textsuperscript{\rm 2}Massachusetts Institute of Technology \\
    \textsuperscript{\rm 3}Amazon \\
}

\usepackage{bibentry}

\begin{document}

\maketitle

\begin{abstract}
Users on e-commerce platforms can be uncertain about their preferences early in their search. Queries to recommendation systems are frequently ambiguous, incomplete, or weakly specified. Agentic systems are expected to proactively reason, ask clarifying questions, and act on the user’s behalf, which makes handling such ambiguity increasingly important. In existing platforms, ambiguity led to excessive interactions and question fatigue or overconfident recommendations prematurely collapsing the search space. 
We present an Interactive Decision Support System (IDSS) that addresses ambiguous user queries using entropy as a unifying signal. IDSS maintains a dynamically filtered candidate product set and quantifies uncertainty over item attributes using entropy. This uncertainty guides adaptive preference elicitation by selecting follow-up questions that maximize expected information gain. When preferences remain incomplete, IDSS explicitly incorporates residual uncertainty into downstream recommendations through uncertainty-aware ranking and entropy-based diversification, rather than forcing premature resolution.
We evaluate IDSS using review-driven simulated users grounded in real user reviews, enabling a controlled study of diverse shopping behaviors. Our evaluation measures both interaction efficiency and recommendation quality. Results show that entropy-guided elicitation reduces unnecessary follow-up questions, while uncertainty-aware ranking and presentation yield more informative, diverse, and transparent recommendation sets under ambiguous intent. These findings demonstrate that entropy-guided reasoning provides an effective foundation for agentic recommendation systems operating under uncertainty.

\end{abstract}

%

\section{Introduction}

Users interacting with e-commerce recommendation systems can have uncertainty about their preferences, particularly early in the search process. Initial queries may express high-level goals such as budget constraints, general use cases, or vague notions of quality or style, without specifying concrete attribute values \cite{Christakopoulou2016ConversationalRecommender}. This ambiguity creates a fundamental problem: systems must operate under incomplete information while still helping users navigate large product spaces effectively.

Interactive and conversational recommendation approaches provide a natural mechanism for addressing ambiguity by allowing systems to ask clarifying questions and adapt recommendations over multiple turns \cite{Christakopoulou2016ConversationalRecommender}. However, ambiguity cannot always be fully resolved through interaction alone. Excessive follow-up questions can lead to user fatigue, while prematurely committing to a narrow interpretation of user intent can collapse the search space and exclude viable alternatives. At the same time, many existing systems treat preference elicitation, ranking, and result presentation as separate components, which limits their ability to reason consistently about uncertainty as it propagates through the recommendation pipeline.

In this work, we introduce an Interactive Decision Support System (\textbf{IDSS}) designed to handle ambiguous user queries by reasoning explicitly about uncertainty. IDSS adopts an information-theoretic perspective in which entropy over the feasible candidate set quantifies uncertainty about user preferences. Unlike prior conversational systems that primarily use uncertainty to decide which questions to ask, IDSS propagates this uncertainty signal throughout elicitation, ranking, and presentation. This enables the system to both gather information when it is valuable and leverage residual uncertainty when preferences remain incomplete.

IDSS integrates three components within a unified framework. First, it uses entropy to guide preference elicitation, selecting follow-up questions that are informative given the distribution of available options rather than relying on fixed scripts. Second, it adapts ranking strategies based on the degree of remaining uncertainty, combining semantic relevance with explicit consideration of risk along unspecified attributes. This extends classic diversity-aware reranking approaches such as MMR \cite{carbonell1998use} by tying diversification directly to unresolved preference uncertainty rather than treating it as a purely downstream heuristic. Third, IDSS presents recommendations in a structured, diversity-aware format that exposes trade-offs along high-uncertainty dimensions, supporting preference discovery through comparison rather than additional dialogue.

We evaluate IDSS in a car recommendation domain, which provides a realistic and challenging testbed due to its high-dimensional attributes, strong trade-offs, and frequent user uncertainty. This domain has been widely used in prior work on interactive and conversational recommendation \cite{Christakopoulou2016ConversationalRecommender}, making it suitable for comparison. In addition to cars, we briefly applied IDSS to an electronics domain with similar attribute structure and observed qualitatively consistent behavior, suggesting that the framework generalizes beyond a single product category. Evaluation is conducted using review-driven simulated users grounded in real user reviews, following established simulation-based methodologies \cite{Ie2019RecSim}. Results show that entropy-guided elicitation reduces unnecessary questioning, while uncertainty-aware ranking and presentation produce more informative, diverse, and transparent recommendation sets under ambiguous user intent.

Our contributions are threefold. First, we introduce an Interactive Decision Support System that treats uncertainty as a first-class signal shared across elicitation, ranking, and presentation, rather than optimizing these stages independently. Second, we propose entropy-aware ranking and diversification strategies that extend existing diversity-aware methods by explicitly accounting for incomplete preferences. Third, we present an evaluation framework that quantifies the trade-off between interaction burden and recommendation quality under ambiguous queries, providing insight into how interactive design choices affect user outcomes.

\begin{figure*}[t]
  \centering
  \includegraphics[width=0.9\textwidth]{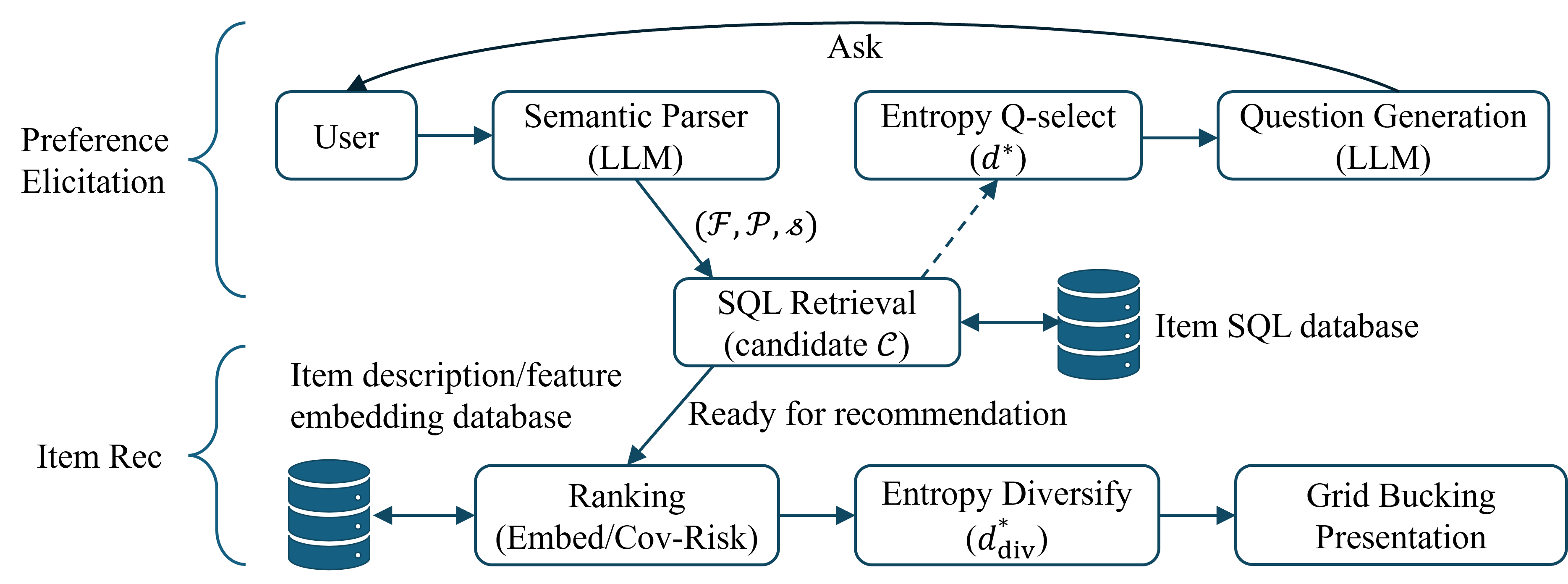}
  \caption{IDSS system overview. The conversational loop uses LLM parsing and entropy-guided question selection, followed by candidate ranking
  and entropy-based grid diversification.}
  \label{fig:architecture}
\end{figure*}

\section{Related Work}
Conversational recommender systems (CRS) have long investigated preference elicitation and adaptive recommendation through multi-turn interaction \cite{Christakopoulou2016ConversationalRecommender}.
With the rise of large language models (LLMs), CRS is increasingly treated as an agentic workflow where an LLM plans and invokes tools such as retrieval, scoring, and memory.
InteRecAgent integrates an LLM with recommender models as tools to support interactive recommendation with explanations, reflection, and modular tool use \cite{Huang2023RecommenderAIAgent}, while RecMind studies an LLM-powered autonomous recommender agent with explicit planning for zero-shot recommendation \cite{Wang2023RecMind}.
Our system contributes to this line of work by combining preference extraction and tool invocation with a controllable, auditable pipeline: the agent translates user intents into explicit SQL constraints over a car database, then invokes embedding-based ranking and reranking modules to produce transparent, diverse shortlists.

A second thread concerns scalable retrieval/ranking and diversity-aware recommendation.
Embedding-based retrieval is a standard approach for candidate generation in production search and recommendation settings \cite{Huang2020EmbeddingFacebookSearch}.
Downstream reranking is often where systems incorporate objectives beyond relevance; surveys highlight the breadth of approaches for fairness and diversity in recommender systems \cite{Zhao2023FairnessDiversitySurvey}.
Recent work also explores using LLMs as flexible rerankers that can reason over multiple criteria when re-ordering candidates \cite{Gao2024LLM4Rerank}.
Building on these ideas, we treat diversification as a design goal for car recommendations, combining structured filtering with semantic ranking and a reranking stage that can explicitly trade off match quality against coverage over relevant attributes, mitigating option duplication.

Finally, we build on evaluation methodologies that use simulation to study how well our recommendation system works with simulated and real users.
Simulation platforms such as RecSim \cite{Ie2019RecSim}, domain-grounded simulators such as KuaiSim \cite{Zhao2023KuaiSim}, and Reinforcement Learning environments with simulated human behavior such as SUBER \cite{Corecco2024SUBER} enable offline experimentation with interactive policies.
More recent approaches use LLMs to generate natural, persona-based user simulators for recommender evaluation \cite{Bougie2025SimUSER,Chen2025RecUserSim,Zhang2025LLMPoweredUserSimulator,Wang2025UserBehaviorSimulationAgents}, while complementary analyses highlight reliability limits and validation challenges for LLM-based simulators \cite{Yoon2024EvaluatingLLMUserSimulators,Zhu2024HowReliableSimulator}.
Our experiments aim to measure the tradeoff between helping users quickly identify good matches and the interaction burden imposed by long dialogues or overly large (or insufficiently diverse) recommendation sets, using controlled variations in interaction design and shortlist size under diverse simulated personas.

\section{Method}
\label{sec:method}

We present IDSS (Interactive Decision Support System), a conversational recommendation framework that combines
(i) entropy-guided preference elicitation and (ii) diversified recommendation presentation.
Figure~\ref{fig:architecture} illustrates the end-to-end architecture: each turn IDSS parses user input into structured
state, retrieves a candidate set under current constraints, optionally asks an informative follow-up question, and
ultimately ranks and presents recommendations in an exploration-friendly format.


\subsection{System}
\label{sec:system}

IDSS is guided by three design principles, each realized by a concrete module in Fig.~\ref{fig:architecture} and detailed in the subsections that follow.
\begin{enumerate}
  \item \textbf{Data-Grounded Elicitation} 
  At each turn, the system asks about the attribute that is most informative \emph{given the currently feasible candidates}.
  Concretely, after filtering the database to obtain a candidate set $\mathcal{C}$, IDSS computes entropy over attribute
  dimensions and selects the next question dimension by maximum entropy (Section~\ref{sec:question_selection}). This ensures
  questions are driven by what is actually available rather than a fixed script.

  \item \textbf{Dual Ranking Strategies} 
  After the interview phase ends, IDSS ranks candidates using one of two complementary strategies depending on the level of
  preference uncertainty: (i) \emph{Embedding Similarity with MMR} to match user intent while avoiding redundancy, and
  (ii) \emph{Coverage-Risk Optimization} to explicitly trade off alignment with stated preferences vs.\ risk from unspecified
  high-uncertainty dimensions (Section~\ref{sec:ranking}).

  \item \textbf{Exploration-Enabling Presentation} 
  Rather than returning a flat list, IDSS groups recommendations along a high-entropy \emph{unspecified} dimension. This makes
  trade-offs visible (e.g., hybrid vs.\ electric) and supports preference discovery through comparison (Section~\ref{sec:diversification}).
\end{enumerate}

These modules are unified by an information-theoretic view: entropy quantifies uncertainty in the candidate space, guiding what to ask (elicitation), how to rank under incomplete information (ranking), and how to present results to encourage exploration (presentation).

\paragraph{Turn-level loop.}
Each turn: (1) parse user input into structured state (Section~\ref{sec:parsing}); (2) retrieve candidates under current filters;
(3) either ask an entropy-selected question (Section~\ref{sec:question_selection}) or stop if further questions are low-value or the user is impatient;
(4) rank candidates (Section~\ref{sec:ranking}); (5) diversify presentation into a grid (Section~\ref{sec:diversification}), with edge cases handled as in
Section~\ref{sec:edge_cases}.

\subsection{Semantic Parsing}
\label{sec:parsing}

Semantic parsing is the entry point of the loop in Fig.~\ref{fig:architecture}: it converts free-form user input into the structured state
used by retrieval, entropy computation, and ranking.

\begin{figure}[H]
  \centering
  \includegraphics[width=\columnwidth]{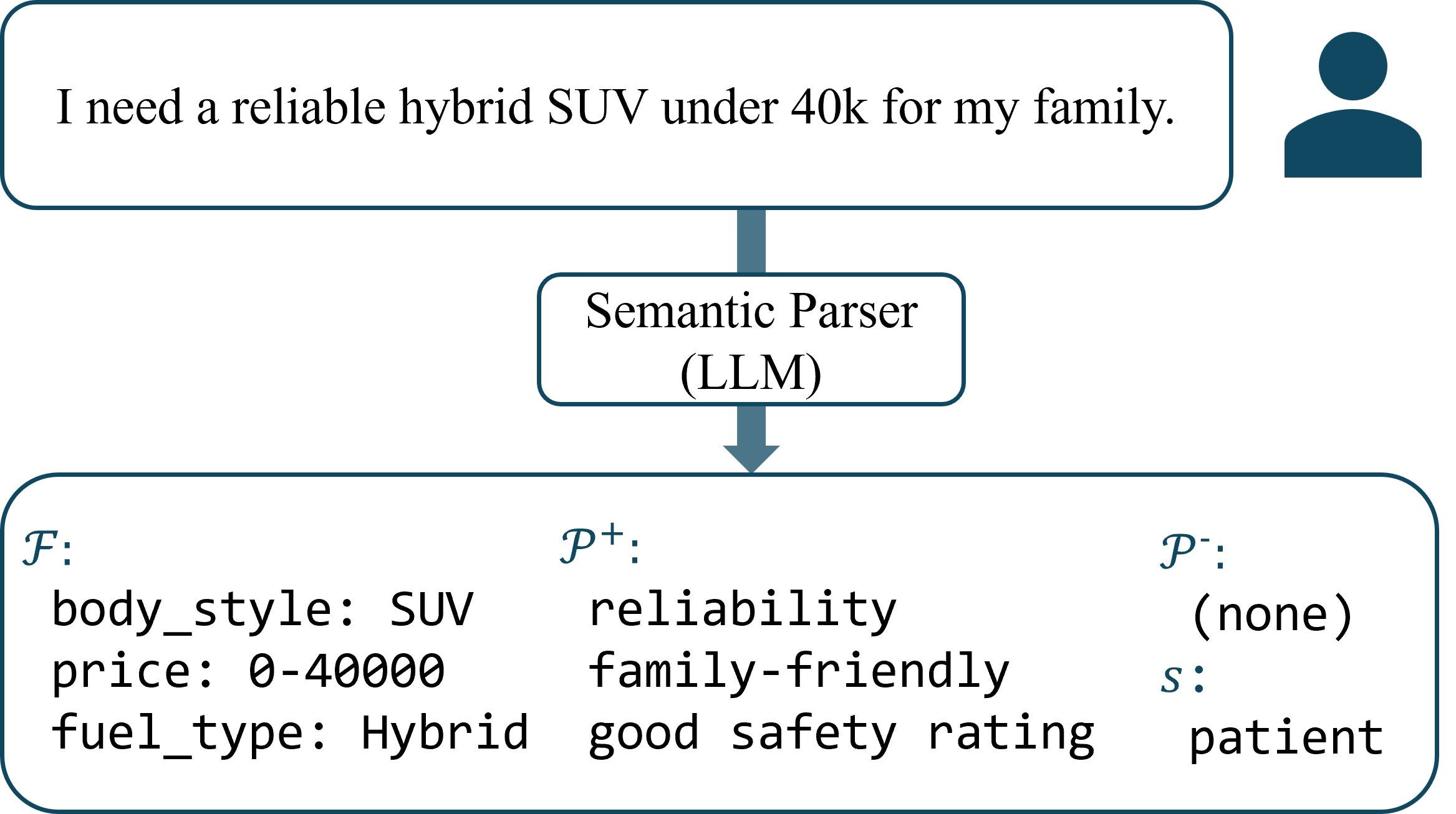}
  \caption{Example of LLM-based semantic parsing that maps a free-form user query into structured hard constraints ($\mathcal{F}$), soft preference cues ($\mathcal{P}^+$), user disliked features ($\mathcal{P}^-$), and user's patience level ($s$) which are subsequently used for downstream retrieval and ranking.}
  \label{fig:semantic_parser}
\end{figure}

An example is illustrated in Fig. \ref{fig:semantic_parser}. Given user query $u$, the semantic parser extracts: 
\begin{equation}
    \text{Parse}(u) \rightarrow (\mathcal{F}, \mathcal{P}, s)
\end{equation}
where $\mathcal{F} = \{f_1, \ldots, f_m\}$ denotes explicit filters (e.g., database attributes), $\mathcal{P} = (\mathcal{P}^+, \mathcal{P}^-)$ captures
implicit preferences (liked and disliked features), and $s \in \{\texttt{patient}, \texttt{impatient}\}$ indicates user engagement level.

We implement the parser using GPT-5 with JSON schema enforcement, ensuring consistent structured outputs. Filters are merged across conversation turns, with newer values overriding previous ones for the same attribute.  The parser also flags impatience signals, detected via prompt instructions that recognize terse responses or explicit skip requests, to enable the system to bypass remaining questions and proceed directly to the recommendation phase. 

\subsection{Entropy-Guided Question Selection}
\label{sec:question_selection}

A central challenge in conversational recommendation is determining \textit{what to ask}. IDSS selects question dimensions using an entropy-guided rule computed over the
current candidate set.

\subsubsection{Entropy Computation}

Let $\mathcal{C}$ denote the candidate set retrieved using current filters $\mathcal{F}$. Let $\mathcal{D}$ be the set of all attribute dimensions.
For each dimension $d \in \mathcal{D}$, we compute Shannon entropy~\cite{shannon1948mathematical}:
\begin{equation}
    H(d) = -\sum_{v \in \text{Val}(d)} p(v) \log_2 p(v)
    \label{eq:entropy}
\end{equation}
where $\text{Val}(d)$ is the set of distinct values for dimension $d$, and
$p(v) = |\{c \in \mathcal{C} : c_d = v\}| / |\mathcal{C}|$ is the proportion of candidates with value $v$.

High entropy indicates that candidates are spread across many values for that dimension, suggesting that user input would effectively partition the candidate space.
Conversely, low entropy (e.g., 95\% of candidates share the same transmission type) indicates the dimension provides limited discriminative value.

For continuous attributes (price, mileage, year), we apply quantile-based discretization into $k=3$ equal-frequency bins before computing entropy. To ensure comparability across dimensions with different cardinalities, we use normalized entropy:
\begin{equation}
  H_{\text{norm}}(d) = \frac{H(d)}{\log_2 |\text{Val}(d)|}
  \label{eq:entropy_norm}
\end{equation}
where $|\text{Val}(d)|$ is the number of distinct values observed for dimension $d$ in the current candidate set (or $k$ for discretized continuous attributes).
This yields values in $[0, 1]$, where 1 indicates a uniform distribution and 0 indicates all candidates share the same value. 

\subsubsection{Dynamic Question Dimension Selection} 


We define three dimension sets:
(i) $\mathcal{D}_{\text{spec}}$ dimensions already specified by explicit filters $\mathcal{F}$,
(ii) $\mathcal{D}_{\text{asked}}$ dimensions previously queried in the dialogue, and
(iii) $\mathcal{D}_{\text{avail}} = \mathcal{D} \setminus (\mathcal{D}_{\text{spec}} \cup \mathcal{D}_{\text{asked}})$.

We select the next question dimension as:
\begin{equation}
    d^* = \underset{d \in \mathcal{D}_{\text{avail}}}{\arg\max} \ H(d).
    \label{eq:dim_selection}
\end{equation}

To avoid low-value questions, we apply a minimum entropy threshold $\tau_H = 0.3$. If no available dimension exceeds $\tau_H$,
the system proceeds directly to recommendation.

\subsubsection{Hybrid Question Generation}

Once $d^*$ is selected, we use an LLM to generate a natural, contextual question:
\begin{equation}
    q = \text{LLM}(d^*, \text{Context}(d^*, \mathcal{C}), \mathcal{H})
\end{equation}
where $\text{Context}(d^*, \mathcal{C})$ provides distribution statistics (e.g., ``40\% gasoline, 35\% hybrid, 25\% electric'') and $\mathcal{H}$ is the conversation history.
This hybrid approach separates \emph{data-driven} dimension selection from \emph{language generation}. Algorithm~\ref{alg:entropy_question} summarizes the selection procedure.

\begin{algorithm}[t]
\caption{Entropy-Guided Question Selection}
\label{alg:entropy_question}
\begin{algorithmic}[1]
\REQUIRE Candidate set $\mathcal{C}$, already specified dimensions $\mathcal{D}_{\text{spec}}$, asked dimensions $\mathcal{D}_{\text{asked}}$, threshold $\tau_H$
\ENSURE Question dimension $d^*$ or $\emptyset$
\STATE $\mathcal{D}_{\text{avail}} \leftarrow \mathcal{D} \setminus (\mathcal{D}_{\text{spec}} \cup \mathcal{D}_{\text{asked}})$
\FOR{each $d \in \mathcal{D}_{\text{avail}}$}
    \STATE Compute $H(d)$ using Eq.~\ref{eq:entropy}
\ENDFOR
\STATE $d^* \leftarrow \arg\max_{d \in \mathcal{D}_{\text{avail}}} H(d)$
\IF{$H(d^*) < \tau_H$}
    \RETURN $\emptyset$
\ENDIF
\RETURN $d^*$
\end{algorithmic}
\end{algorithm}

\subsection{Candidate Ranking}
\label{sec:ranking}

After the interview phase concludes (either $k$ questions asked, entropy falls below threshold, or the user signals impatience),
IDSS retrieves and ranks candidates. We implement two complementary ranking strategies (System principle \#2).

\subsubsection{Embedding Similarity with MMR}

This approach ranks candidates by semantic similarity to a query constructed from extracted preferences, with diversity enforced via
Maximal Marginal Relevance (MMR)~\cite{carbonell1998use}.

We construct a textual query $q_{\text{text}}$ from filters $\mathcal{F}$ and preferences $\mathcal{P}$, then compute similarity using a sentence encoder $\phi$:
\begin{equation}
    \text{sim}(q, c) = \cos(\phi(q_{\text{text}}), \phi(\text{desc}(c)))
\end{equation}
where $\text{desc}(c)$ is a textual description of candidate $c$.

To prevent redundant recommendations, we apply MMR during selection: 
\begin{equation}
    \text{MMR}(c) = \lambda \cdot \text{sim}(q, c) - (1-\lambda) \cdot \max_{c' \in \mathcal{S}} \text{sim}(c, c')
    \label{eq:mmr}
\end{equation}
where $\mathcal{S}$ is the set of already-selected candidates and $\lambda$ controls the relevance-diversity trade-off. We set $\lambda = 0.85$, which prioritizes relevance while penalizing near-duplicate recommendations. This value was selected empirically to balance precision with intra-list diversity.

\subsubsection{Coverage-Risk Optimization}


This approach ranks candidates by maximizing alignment with liked features while penalizing alignment with disliked features, using phrase-level semantic matching against item reviews.

For each item $v$, we extract pros and cons phrases from aggregated reviews and pre-compute their embeddings using a sentence transformer (all-mpnet-base-v2).
Given a user's liked feature $j \in \mathcal{P}^+$ with embedding $\mathbf{e}_j$, we compute alignment by matching against item $v$'s pros phrases $\{z_{v,1}, \ldots, z_{v,K}\}$:
\begin{equation}
  \text{Pos}_j(v) = \max_{k} \, \varphi\bigl(\cos(\mathbf{e}_j, \mathbf{z}_{v,k})\bigr)
\end{equation}
where $\varphi(t) = \max(0, t - \tau)$ is a threshold function that filters weak matches.
We set $\tau = 0.6$ to ensure only confident matches contribute to the score.

Similarly, for disliked features $r \in \mathcal{P}^-$, we compute risk alignment against cons phrases:
\begin{equation}
  \text{Neg}_r(v) = \max_{k} \, \varphi\bigl(\cos(\mathbf{e}_r, \mathbf{z}_{v,k})\bigr)
\end{equation}

We select a set $\mathcal{S}$ of $K$ items by greedy maximization of:
\begin{equation}
  \max_{|\mathcal{S}|=K} \underbrace{\sum_{j \in \mathcal{P}^+} \max_{v \in \mathcal{S}} \text{Pos}_j(v)}_{\text{Coverage}}
  - \lambda \underbrace{\sum_{r \in \mathcal{P}^-} \max_{v \in \mathcal{S}} \text{Neg}_r(v)}_{\text{Risk}}
  \label{eq:coverage_risk}
\end{equation}
The coverage term rewards sets that collectively satisfy all liked features, while the risk term penalizes sets containing items with disliked attributes. We set $\lambda = 0.5$ to balance these objectives; this value was selected via grid search on a validation set, with results robust across $\lambda \in [0.4, 0.7]$.

We apply a greedy algorithm that iteratively selects the item with highest marginal gain:
\begin{equation}
  v^* = \arg\max_{v \notin \mathcal{S}} \Bigl[ \Delta\text{Cov}(v | \mathcal{S}) - \lambda \cdot \Delta\text{Risk}(v | \mathcal{S}) \Bigr]
\end{equation}
where $\Delta\text{Cov}(v | \mathcal{S})$ is the marginal coverage gain from adding $v$ to $\mathcal{S}$.
Due to submodularity of the coverage term, this greedy approach provides a $(1 - 1/e)$ approximation guarantee.

\subsection{Entropy-Based Result Diversification}
\label{sec:diversification}

Rather than presenting a flat ranked list, IDSS organizes results into a grid structure that facilitates systematic exploration
(System principle \#3).

\subsubsection{Diversification Dimension Selection}

After ranking, we select a presentation dimension to organize the top-$K$ results.
Let $\mathcal{C}_{\text{ranked}}$ denote the ranked candidate set output by the ranking methods. We apply the normalized entropy criterion (Eq.~\ref{eq:entropy_norm}) restricted to unspecified dimensions:
\begin{equation}
  d_{\text{div}} = \underset{d \in \mathcal{D}_{\text{unspec}}}{\arg\max} \ H_{\text{norm}}(d \mid \mathcal{C}_{\text{ranked}})
\end{equation}
where $\mathcal{D}_{\text{unspec}} = \mathcal{D} \setminus \mathcal{D}_{\text{spec}}$ excludes dimensions already constrained by user filters.
The intuition is that users benefit from seeing variety along dimensions they have not committed to, enabling preference discovery through comparison.


\subsubsection{Grid Construction}

Given $d_{\text{div}}$ and target grid dimensions $r \times n$ (where $r$ is the number of rows and $n$ is items per row), we partition candidates by their value
on $d_{\text{div}}$, then select top-ranked items from each partition. Algorithm~\ref{alg:bucketing} describes this process.

\begin{algorithm}[t]
\caption{Entropy-Based Bucketing}
\label{alg:bucketing}
\begin{algorithmic}[1]
\REQUIRE Ranked candidates $\mathcal{C}$, dimension $d_{\text{div}}$, grid size $r \times n$
\ENSURE Recommendation grid $\mathcal{G}$
\STATE $\mathcal{B} \leftarrow \text{Partition}(\mathcal{C}, d_{\text{div}})$ \COMMENT{Group by value on $d_{\text{div}}$}
\STATE Sort $\mathcal{B}$ by partition size (descending)
\STATE $\mathcal{G} \leftarrow []$
\FOR{$i = 1$ to $\min(r, |\mathcal{B}|)$}
  \STATE $\text{row}_i \leftarrow \text{Top}_n(\mathcal{B}_i)$ \COMMENT{Select $n$ highest-ranked items from partition $i$}
  \STATE $\mathcal{G}.\text{append}(\text{row}_i, \text{Label}(\mathcal{B}_i))$
\ENDFOR
\RETURN $\mathcal{G}$
\end{algorithmic}
\end{algorithm}

Here $\text{Top}_n(\mathcal{B}_i)$ returns the $n$ highest-ranked items within partition $\mathcal{B}_i$ (preserving the ranking from the previous stage), and
$\text{Label}(\mathcal{B}_i)$ returns a human-readable description of the partition (e.g., ``Hybrid'' for fuel type, or ``\$20K--\$30K'' for price). Each row thus represents a distinct value along $d_{\text{div}}$, helping users understand the trade-off space through direct comparison. 

\subsection{Handling Edge Cases}
\label{sec:edge_cases}

\subsubsection{Zero-Result Queries}

When filters are overly restrictive and $\mathcal{C}=\emptyset$, IDSS applies progressive filter relaxation. Filters are ranked by importance (cosmetic attributes relaxed first, fundamental requirements last), and iteratively removed until results are found. The system tracks relaxed filters to inform users which criteria could not be fully satisfied.

\subsubsection{Impatience Detection}

The LLM-based semantic parser detects signals of user impatience through user input. When detected, the system terminates the interview early and proceeds directly to recommendation, respecting user agency over strict adherence to the configured $k$ value.

\section{Evaluation}
\label{sec:eval}

We evaluate our conversational car recommendation system using review-driven simulated users. The evaluation is designed to jointly measure (i) dialogue efficiency and elicitation quality (how the agent asks follow-up questions) and (ii) recommendation quality and diversity (how well the final shortlist matches the user while avoiding redundant options).

\subsection{Review-driven simulated users}
\begin{figure*}[t]
  \centering
  \includegraphics[width=0.9\textwidth]{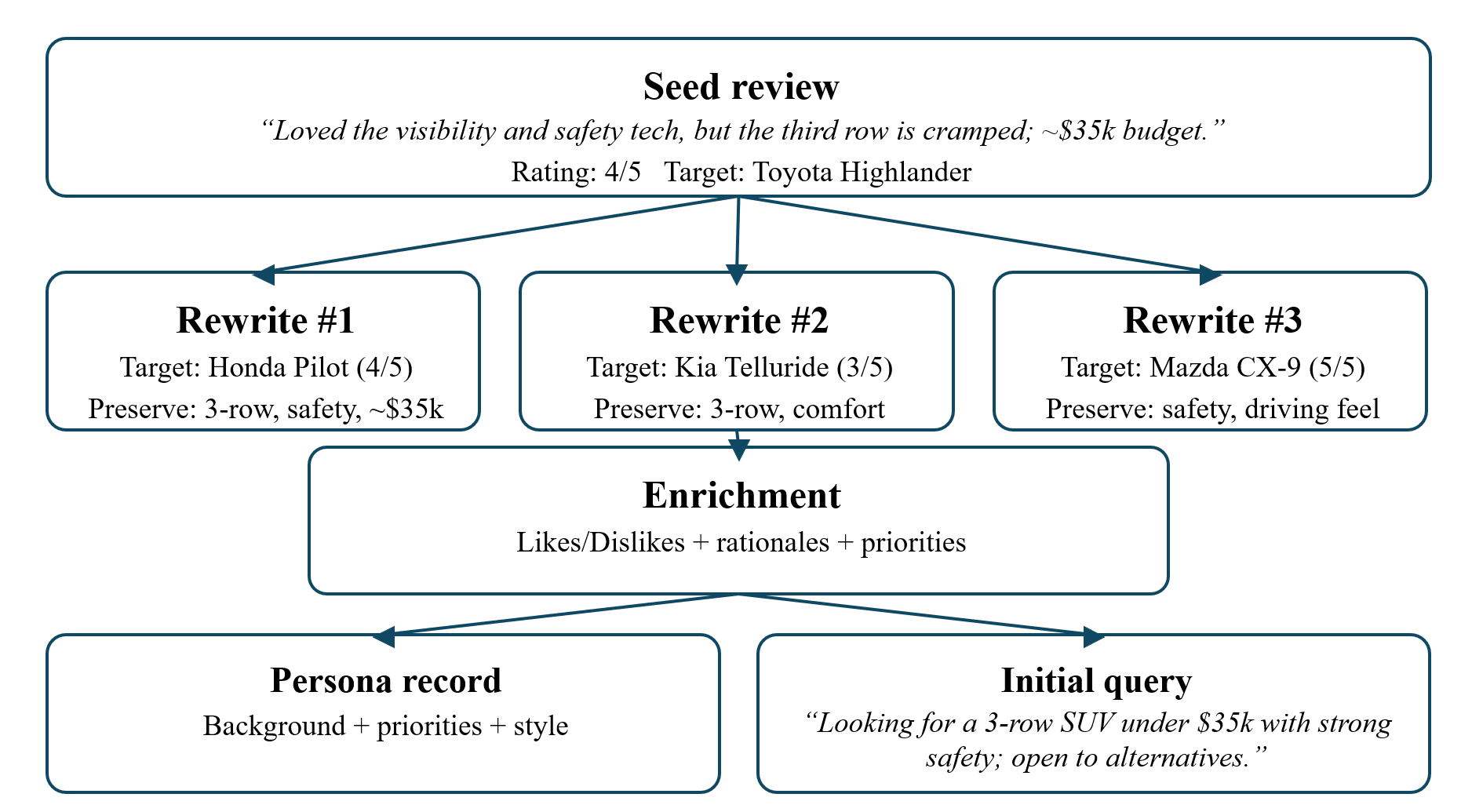}
  \caption{Example of review-driven simulation. A single seed review is rewritten into multiple stylistically consistent variants targeting different vehicles while preserving constraints and priorities. Each review is enriched into a structured record, which yields a persona and an initial query for the interactive evaluation.}
  \label{fig:seed_to_persona}
\end{figure*}

\paragraph{Seed reviews and augmentation.}
We begin with real car reviews from online sources and use them as the behavioral anchor for simulated shoppers. To increase coverage over makes/models beyond the raw review distribution, we generate additional review variants by rewriting each seed review to describe alternative target vehicles while preserving the original sentiment, tone, and rating context. This augmentation yields multiple stylistically consistent reviews for diverse make/model pairs. An example is illustrated in Fig. \ref{fig:seed_to_persona}.

\paragraph{Persona enrichment.}
Each review is then converted into a structured persona record via LLM-based information extraction. The extracted persona includes: (i) explicit preferences over vehicle configurations with rationales, (ii) a brief shopping intention statement, and (iii) aggregated preference signals such as mentioned makes/models/years, preference for new vs.\ used, body style, fuel type, openness to alternatives, and other priorities (e.g., safety, reliability, comfort, budget).

\paragraph{Persona turns and query generation.}
From the enriched persona record, we generate a realistic user turn consisting of a concise initial query plus latent behavioral attributes: writing style, interaction style, family background, goal, and an estimated maximum price based on the preferences.

\subsection{Simulation protocol}
For each persona turn, we run a multi-turn interaction between:
(1) the recommendation agent, which may ask follow-up questions and ultimately returns a list of recommended vehicles,
(2) a simulated user that answers follow-up questions in a consistent style, and
(3) an automatic judge that scores both follow-up questions and recommendations.

The interaction starts from the persona's initial query. If the agent asks a follow-up question, the simulator responds. We vary key interaction parameters to quantify the tradeoff between information-gathering effort and recommendation quality.

\subsection{Follow-up question evaluation}
Each follow-up question is evaluated by an LLM judge along two dimensions:
\textbf{Relevance} (does the question directly relate to the user’s stated query and inferred goal?) and
\textbf{Newness} (does it ask for information not already requested earlier in the dialogue?).
The judge returns binary pass/fail labels for each dimension along with a confidence score and short rationale.

\subsection{Recommendation evaluation}
Given the final recommendation list, an LLM judge evaluates each recommended vehicle against the persona's query and the full conversation history. The judge produces:
(i) an overall satisfaction label (satisfied/unsatisfied),
(ii) a confidence score in $[0,1]$,
and (iii) attribute-level assessments (satisfied / not satisfied / not mentioned) with brief rationales for core facets such as price, year, make, model, body type, fuel type, condition, and other constraints.

\paragraph{Top-$k$ metrics.}
Let $\text{rel}_i \in \{0,1\}$ denote whether the $i$-th recommended vehicle is judged satisfactory.
We report:
\begin{itemize}
    \item \textbf{SatisfiedCount@k}: the number of satisfactory vehicles among the top $k$
    \item \textbf{Precision@k}: $\frac{1}{k}\sum_{i=1}^{k}\text{rel}_i$
    \item \textbf{nDCG@k} with binary relevance: $
\mathrm{DCG}@k = \sum_{i=1}^{k} \frac{2^{\text{rel}_i}-1}{\log_2(i+1)},\;
\mathrm{nDCG}@k = \mathrm{DCG}@k / \mathrm{IDCG}@k
$
\end{itemize}
\paragraph{Judge confidence.}
For each judged decision,
the judge returns a \emph{confidence score} $c \in [0,1]$ together with a short rationale.
Operationally, $c$ is produced by the judge model as a calibrated self-assessment of how strongly the
evidence in the conversation supports its binary label.
We use confidence in two ways: (i) to report confidence-filtered metrics that emphasize robust wins,
and (ii) to trigger reassessment when the judge appears uncertain.

\paragraph{High Certainty Confidence Threshold.}
Because LLM judges can be noisy on borderline cases,
we set a confidence threshold $\tau$ and treat judgments with $c<\tau$ as \emph{ambiguous}.
Our default $\tau=0.51$ is intentionally close to a ``bare majority'' cutoff: it minimizes false discards when the judge
is only slightly unsure, while still separating clearly supported decisions ($c>\tau$) from near-coin-flips ($c\approx 0.5$).
Concretely, this choice makes the filtered metrics sensitive to strong preference matches without overly shrinking the
effective sample size.

\paragraph{Robustness via reassessment.}
When the average confidence across the recommendation list is low, we rerun the recommendation assessment multiple times
and aggregate labels via majority vote (breaking ties using confidence).
This yields stable estimates under judge variability while keeping the protocol fully offline and reproducible.
We report both raw metrics and confidence-filtered metrics (items with $c\ge\tau$) to transparently separate
high certainty outcomes from ambiguous cases.

\begin{table*}[h!]
\centering
\small
\begin{tabular}{cllcccc}
\toprule
\textbf{Query Data Type} & \textbf{Method} & \textbf{Config} & \textbf{Prec@9} $\uparrow$ & \textbf{NDCG@9} $\uparrow$ & \textbf{Sat@9} $\uparrow$ & \textbf{ILD} $\uparrow$ \\
\midrule
\multirow{7}{*}{Short}
& \multirow{4}{*}{ES}
& Full & \underline{.903} {\scriptsize $\pm$.025} & \underline{.941} {\scriptsize $\pm$.011} & \underline{.772} {\scriptsize $\pm$.032} & \textbf{.779} {\scriptsize $\pm$.043} \\
& & $-$MMR & \textbf{.918} {\scriptsize $\pm$.020} & \textbf{.952} {\scriptsize $\pm$.007} & .764 {\scriptsize $\pm$.019} & .279 {\scriptsize $\pm$.011} \\
& & $-$EntropyQ & .880 {\scriptsize $\pm$.056} & .927 {\scriptsize $\pm$.032} & .732 {\scriptsize $\pm$.051} & \underline{.611} {\scriptsize $\pm$.015} \\
& & $-$MMR and EntropyQ & .902 {\scriptsize $\pm$.019} & .931 {\scriptsize $\pm$.023} & \textbf{.778} {\scriptsize $\pm$.016} & .231 {\scriptsize $\pm$.004} \\
\cmidrule{2-7}
& \multirow{4}{*}{CR}
& Full & \textbf{.887} {\scriptsize $\pm$.025} & \underline{.900} {\scriptsize $\pm$.025} & \textbf{.758} {\scriptsize $\pm$.037} & \underline{.319} {\scriptsize $\pm$.008} \\
& & $-$MMR & \underline{.883} {\scriptsize $\pm$.012} & \textbf{.920} {\scriptsize $\pm$.005} & .719 {\scriptsize $\pm$.024} & .179 {\scriptsize $\pm$.003} \\
& & $-$EntropyQ & .848 {\scriptsize $\pm$.046} & .882 {\scriptsize $\pm$.045} & .706 {\scriptsize $\pm$.034} & \textbf{.327} {\scriptsize $\pm$.006} \\
& & $-$MMR and EntropyQ & .866 {\scriptsize $\pm$.028} & .893 {\scriptsize $\pm$.026} & \underline{.751} {\scriptsize $\pm$.025} & .217 {\scriptsize $\pm$.020} \\
\midrule
\multirow{7}{*}{Long}
& \multirow{4}{*}{ES}
& Full & \underline{.744} {\scriptsize $\pm$.036} & \underline{.868} {\scriptsize $\pm$.037} & \underline{.459} {\scriptsize $\pm$.095} & \textbf{.412} {\scriptsize $\pm$.047} \\
& & $-$MMR & \textbf{.837} {\scriptsize $\pm$.037} & \textbf{.891} {\scriptsize $\pm$.043} & \textbf{.507} {\scriptsize $\pm$.033} & .241 {\scriptsize $\pm$.029} \\
& & $-$EntropyQ & .719 {\scriptsize $\pm$.062} & .779 {\scriptsize $\pm$.053} & .401 {\scriptsize $\pm$.053} & \underline{.405} {\scriptsize $\pm$.055} \\
& & $-$MMR and EntropyQ & .662 {\scriptsize $\pm$.085} & .689 {\scriptsize $\pm$.071} & .417 {\scriptsize $\pm$.050} & .177 {\scriptsize $\pm$.026} \\
\cmidrule{2-7}
& \multirow{4}{*}{CR}
& Full & \underline{.801} {\scriptsize $\pm$.108} & \underline{.875} {\scriptsize $\pm$.106} & \underline{.445} {\scriptsize $\pm$.061} & \underline{.299} {\scriptsize $\pm$.021} \\
& & $-$MMR & \textbf{.863} {\scriptsize $\pm$.033} & \textbf{.909} {\scriptsize $\pm$.024} & \textbf{.504} {\scriptsize $\pm$.015} & .223 {\scriptsize $\pm$.020} \\
& & $-$EntropyQ & .753 {\scriptsize $\pm$.063} & .809 {\scriptsize $\pm$.065} & .406 {\scriptsize $\pm$.043} & \textbf{.306} {\scriptsize $\pm$.024} \\
& & $-$MMR and EntropyQ & .724 {\scriptsize $\pm$.013} & .725 {\scriptsize $\pm$.013} & .431 {\scriptsize $\pm$.008} & .195 {\scriptsize $\pm$.032} \\
\bottomrule
\end{tabular}
\caption{Ablation study of ranking and diversity components under short and long interaction settings. We evaluate the impact of removing MMR-based diversification and entropy-based follow-up questioning (EntropyQ) on recommendation quality for Embedding Similarity (ES) and Coverage-Risk (CR) methods. Results report mean $\pm$ standard deviation over three runs. Bold indicates best and underline indicates second best per section.}
\label{tab:ablation_rec}
\end{table*}

\paragraph{Attribute satisfaction rates.}
For each attribute $a$, we compute an attribute satisfaction rate:
\\\\
$
\mathrm{AttrSat}(a) = \frac{\#\text{ satisfied assessments for }a}{\#\text{ assessed (non-null) for }a}
$
\\\\
which yields interpretable diagnostics on which constraints are most frequently violated.

\section{Experimental Results}
\label{sec:results}

We evaluate IDSS using the review-driven simulation framework described in Section~\ref{sec:eval}.
We consider two query verbosity settings: \textbf{Short} queries (under 10 words, e.g., ``Looking for a used SUV under \$30k'')
and \textbf{Long} queries (under 120 words, including detailed preferences, likes/dislikes, and contextual factors).
The Short setting represents underspecified user requests, while the Long setting captures multi-faceted and explicitly constrained user intents. We constructed a dataset of 150 personas following the procedure described in Section~\ref{sec:eval}, and evaluated IDSS using both ranking methods introduced in Section~\ref{sec:ranking}.

\subsection{Ablation Study: Ranking and Diversification}

Table~\ref{tab:ablation_rec} reports ablation results for the two ranking methods, Embedding Similarity (ES) and Coverage-Risk (CR), under both query settings.
We analyze the effect of removing MMR-based diversification and entropy-guided follow-up question selection (EntropyQ).

\paragraph{Impact of MMR on diversity and relevance.}
Across both query settings and ranking methods, disabling MMR leads to a substantial reduction in intra-list diversity (ILD).
For example, under Short queries with ES, ILD decreases from 0.779 (Full) to 0.279 ($-$MMR), and under Long queries from 0.412 to 0.241.
Similar trends are observed for CR.
These results indicate that MMR is the primary contributor to diversity in the final recommendation list.

At the same time, removing MMR often improves relevance-oriented metrics. For instance, with ES under Long queries, Prec@9 increases from 0.744 (Full) to 0.837 ($-$MMR).
This pattern reflects a standard relevance--diversity trade-off, where diversity-aware re-ranking can slightly reduce top-$k$ relevance while mitigating redundancy.
The Full configuration achieves a balance between these objectives by preserving most of the relevance gains while substantially improving diversity.

\paragraph{Impact of entropy-guided follow-up questions.}
Removing EntropyQ consistently degrades recommendation quality across methods, with a more pronounced effect in the Long query setting.
For ES under Long queries, Prec@9 drops from 0.744 (Full) to 0.719 ($-$EntropyQ), and for CR from 0.801 to 0.753.
These trends suggest that entropy-guided follow-up questions provide useful clarification signals even when users initially express detailed preferences.

\paragraph{Comparison between Short and Long queries.}
Overall recommendation performance is lower under the Long setting
(e.g., ES Full: Prec@9 = 0.903 for Short vs.\ 0.744 for Long).
Although Long queries provide richer information, they also impose more specific and potentially conflicting constraints, making them harder to satisfy. We therefore view the Long setting as a more challenging evaluation regime.

\paragraph{Comparison between ranking methods.}
Across both query settings, Embedding Similarity (ES) and Coverage-Risk (CR) exhibit complementary strengths.
ES consistently performs well under the Short setting, where user queries are brief and underspecified.
In this regime, ranking by semantic similarity provides a robust inductive bias, allowing the system to retrieve plausible candidates even when explicit constraints are sparse or only partially stated.

In contrast, CR shows stronger performance under the Long setting, particularly on precision-oriented metrics.
Long queries contain richer and more explicit constraints, which align naturally with CR’s explicit coverage optimization over satisfied attributes.
By directly optimizing constraint satisfaction rather than relying solely on semantic proximity, CR better exploits detailed preference signals when they are available.

\subsection{Ablation Study: Follow-up Question Selection}

\begin{table}[t]
  \centering
  \small
  \begin{tabular}{llcc}
  \toprule
  \textbf{Data} & \textbf{Config} & \textbf{Relevance} $\uparrow$ & \textbf{Newness} $\uparrow$ \\
  \midrule
  \multirow{2}{*}{Short}
  & w/ EntropyQ & \textbf{1.00} {\scriptsize $\pm$.000} & \textbf{.946} {\scriptsize $\pm$.056} \\
  & w/o EntropyQ & .967 {\scriptsize $\pm$.058} & .602 {\scriptsize $\pm$.047} \\
  \midrule
  \multirow{2}{*}{Long}
  & w/ EntropyQ & \textbf{1.00} {\scriptsize $\pm$.000} & .976 {\scriptsize $\pm$.032} \\
  & w/o EntropyQ & \textbf{1.00} {\scriptsize $\pm$.000} & \textbf{.980} {\scriptsize $\pm$.021} \\
  \bottomrule
  \end{tabular}
  \caption{Ablation study of follow-up question selection under short and long query-verbosity settings. We evaluate entropy-based question selection (EntropyQ) using Relevance and Newness metrics. Results report mean $\pm$ standard deviation over three runs.}
  \label{tab:ablation_question}
  \end{table}
  
Table~\ref{tab:ablation_question} isolates the effect of entropy-guided question selection
on the quality of follow-up questions.

\paragraph{Effect on question novelty.}
EntropyQ substantially improves the novelty of follow-up questions in the Short setting.
Newness increases from 0.602 without EntropyQ to 0.946 with EntropyQ, indicating that entropy-based selection effectively targets attributes with high residual uncertainty
when initial user input is sparse.

\paragraph{Diminishing gains for Long queries.}
In the Long setting, the difference in Newness between configurations narrows (0.976 with EntropyQ vs.\ 0.980 without).
When users specify many preferences upfront, fewer high-uncertainty dimensions remain, limiting the marginal benefit of entropy-guided selection.

\paragraph{Relevance remains consistently high.}
Across all settings and configurations, relevance scores remain near perfect (0.967--1.00),
suggesting that both entropy-guided and baseline strategies generate on-topic follow-up questions.
The primary advantage of EntropyQ lies in eliciting novel information rather than improving topical alignment.

\paragraph{User survey.}
To complement our simulation-based evaluation, we ran a small pilot survey with $n=12$ participants. Each participant issued a natural query to our prototype and compared three interaction policies that ask $0$, $1$, or $2$ follow-up questions before returning a final shortlist. Participants rated how well recommendations matched their intent on a 1 to 7 scale and ranked the three outputs by quality. Ratings increased with more elicitation and $2$ follow-ups was ranked best by $9/12$ participants, while $0$ follow-ups was never ranked best and was ranked worst by $10/12$. Qualitative feedback echoed our findings: participants reported that follow-ups helped surface omitted constraints (e.g., drivetrain or budget), and the diversified presentation made results easier to compare and more ``organized'' (e.g., by highlighting trade-offs such as fuel type or wheel drive).

\section{Conclusion}

We presented IDSS, an interactive decision support system that treats preference uncertainty as a first-class signal throughout conversational recommendation.
By using entropy to guide both preference elicitation and downstream ranking and diversification, IDSS adapts its behavior to varying levels of user specificity. Through review-driven simulation, we showed that entropy-guided questioning improves the novelty of follow-up interactions when initial queries are sparse, while uncertainty-aware ranking and diversification enable a balanced trade-off between relevance and diversity in the final recommendation
set. Our results highlight the importance of jointly designing questioning and ranking strategies, rather than treating them as independent components, for agentic recommendation systems operating under ambiguous user intent.

\section{Limitations and Future Work}

Our evaluation relies on review-driven LLM-based simulated users, which enables controlled and scalable experimentation but may not fully capture the variability and strategic behavior of real users. An important direction for future work is to validate IDSS through human-subject studies, focusing on user satisfaction, interaction efficiency, and perceived transparency.

In addition, our experiments focus on the automotive domain. While the proposed entropy-guided elicitation and uncertainty-aware ranking mechanisms are domain-agnostic, adapting IDSS to other decision-support settings may require re-specifying attributes
and diversification objectives. Future work will explore broader domains and more adaptive strategies that dynamically balance relevance and diversity as user intent evolves during interaction.

\bibliography{aaai2026}

\end{document}